\documentclass{article}
\usepackage{ijcai19}

\usepackage{times}
\usepackage{soul}
\usepackage{url}
\usepackage[hidelinks]{hyperref}
\usepackage[utf8]{inputenc}
\usepackage[small]{caption}

\usepackage{graphicx, subfigure}

\usepackage{amsmath}
\usepackage{amssymb}
\usepackage{booktabs}
\usepackage{algorithm}
\usepackage{algorithmic}

\usepackage{multirow}

\title{Open-Ended Long-Form Video Question Answering via Hierarchical Convolutional Self-Attention Networks}

\author{
Zhu Zhang$^1$
\and
Zhou Zhao\footnote{Zhou Zhao is the corresponding author.}$^1$\and
Zhijie Lin$^1$\and
Jingkuan Song$^2$\And
Xiaofei He$^3$
\affiliations
$^1$College of Computer Science, Zhejiang University, China\\
$^2$University of Electronic Science and Technology of China, China\\
$^3$State Key Lab of CAD\&CG, Zhejiang University, China
\emails
\{zhangzhu, zhaozhou, linzhijie\}@zju.edu.cn,
\{jingkuan.song, xiaofeihe\}@gmail.com
}

\begin{document}

\maketitle

\begin{abstract}
Open-ended video question answering aims to automatically generate the natural-language answer from referenced video contents according to the given question. Currently, most existing approaches focus on short-form video question answering with multi-modal recurrent encoder-decoder networks. Although these works have achieved promising performance, they may still be ineffectively applied to long-form video question answering due to the lack of long-range dependency modeling and the suffering from the heavy computational cost. To tackle these problems, we propose a fast Hierarchical Convolutional Self-Attention encoder-decoder network(HCSA). Concretely, we first develop a hierarchical convolutional self-attention encoder to efficiently model long-form video contents, which builds the hierarchical structure for video sequences and captures question-aware long-range dependencies from video context. We then devise a multi-scale attentive decoder to incorporate multi-layer video representations for answer generation, which avoids the information missing of the top encoder layer. The extensive experiments show the effectiveness and efficiency of our method.
\end{abstract}

\section{Introduction}
Open-ended video question answering is an important problem in visual information retrieval, which automatically generates the accurate answer from referenced video contents according to the given question. Most existing works employ the multi-modal recurrent encoder-decoder framework, which first encodes the multi-modal video and question contents into a joint representation, and then generates the natural-language answer~\cite{xue2017unifying,zhao2018vqa}. Although these approaches have achieved excellent performance in short-form video question answering, they may still be ineffectively applied to the long-form video question answering due to the lack of long-range dependency modeling and the suffering from the heavy computational cost.
\begin{figure}[t]
\centering
\includegraphics[width=0.47\textwidth]{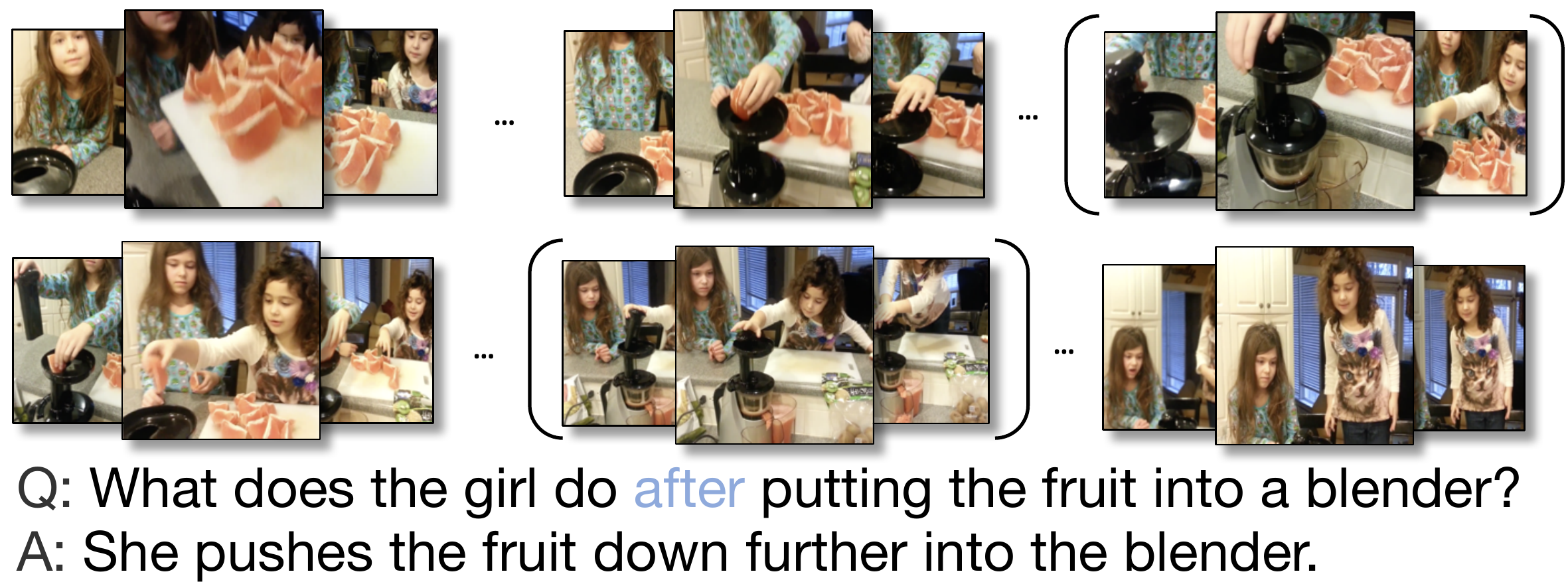}
\caption{Open-Ended Long-Form Video Question Answering.}\label{fig:example}
\end{figure}

The long-form video often contains the complex information of the targeted objects that evolve over time~\cite{krishna2017dense}. We illustrate a concrete example in Figure~\ref{fig:example}. The answer generation to the question \lq\lq What does the girl do after putting the fruit into a blender?\rq\rq\text{ }requires the accurate information extraction from long-form video contents. However, the sequential processing of long-form video contents by recurrent neural networks remains the network depth-in-length, which causes the heavy computational cost and insufficiency of long-range dependency modeling. Therefore, the simple extension of multi-modal recurrent encoder-decoder networks is difficult to generate a satisfactory answer for open-ended long-form video question answering.

Recently, the convolutional sequence-to-sequence method has been applied in fast conditional text generation~\cite{gehring2017convolutional}.
Based on the convolutional sequence modeling, we are able to improve the computational efficiency of long-form video encoding. However, the original convolutional sequence-to-sequence method still has some limitations. Firstly, the temporal convolution keeps the sequential length unchanged, so the multi-layer convolution operation leads to the large memory requirement, especially for long-form video sequences. Secondly, as shown in Figure~\ref{fig:example}, the crucial clues for question answering often are contained in continuous segments rather than discrete frames, but the original convolutional sequence-to-sequence strategy ignores the hierarchical structure of long-form video contents. Thirdly, this method still remains in local modeling and fails to capture long-range dependencies from video context.

We tackle these problems by a hierarchical convolution self-attention encoder, which can efficiently learn multi-layer video semantic representations. Specifically, the encoder consists of multiple convolution self-attention layers, and each layer contains two convolution units, an attentive segmentation unit and a question-aware self-attention unit. The attentive segmentation unit splits long-form video contents into different segments and leverages question information to learn attentive segment-level representations, which builds the hierarchical structure of long-form video contents and explores multi-scale visual clues for question answering. Moreover, this method reduces the video sequential length layer by layer and is helpful to alleviate memory load.
On the other hand, the question-aware self-attention unit exploits the question evidence to capture long-range dependencies from video context. Concretely, different from the conventional self-attention developed for machine translation~\cite{vaswani2017attention}, we consider that a large portion of contents in long-form videos are irrelevant to the given question and even interfere with question answering, thus utilize the question information as guidance for filtering the unnecessary context.

After encoding long-form video contents, we consider the decoder network for answer generation. Previous works~\cite{bahdanau2014neural,anderson2018bottom} often adopt the recurrent neural network with attention mechanism to generate the natural-language answer. Different from them, we develop a multi-scale attentive recurrent decoder to incorporate multi-layer semantic representations from the hierarchical video encoder according to the given question, which avoids the information missing of the top layer.

In summary, the main contributions of this paper are as follows:
\begin{itemize}
\item we develop a hierarchical convolutional self-attention encoder to efficiently explore the hierarchical structure of long-form video contents and capture question-aware long-range dependencies from video context. 
\item We devise a multi-scale attentive decoder to incorporate multi-layer video semantic representations for answer generation, which avoids information missing only from the top layer of the hierarchical encoder.
\item The extensive experiments on a public long-form video question answering dataset validate the effectiveness and efficiency of our method.
\end{itemize}

\begin{figure*}[t]
\centering
\includegraphics[width=0.95\textwidth]{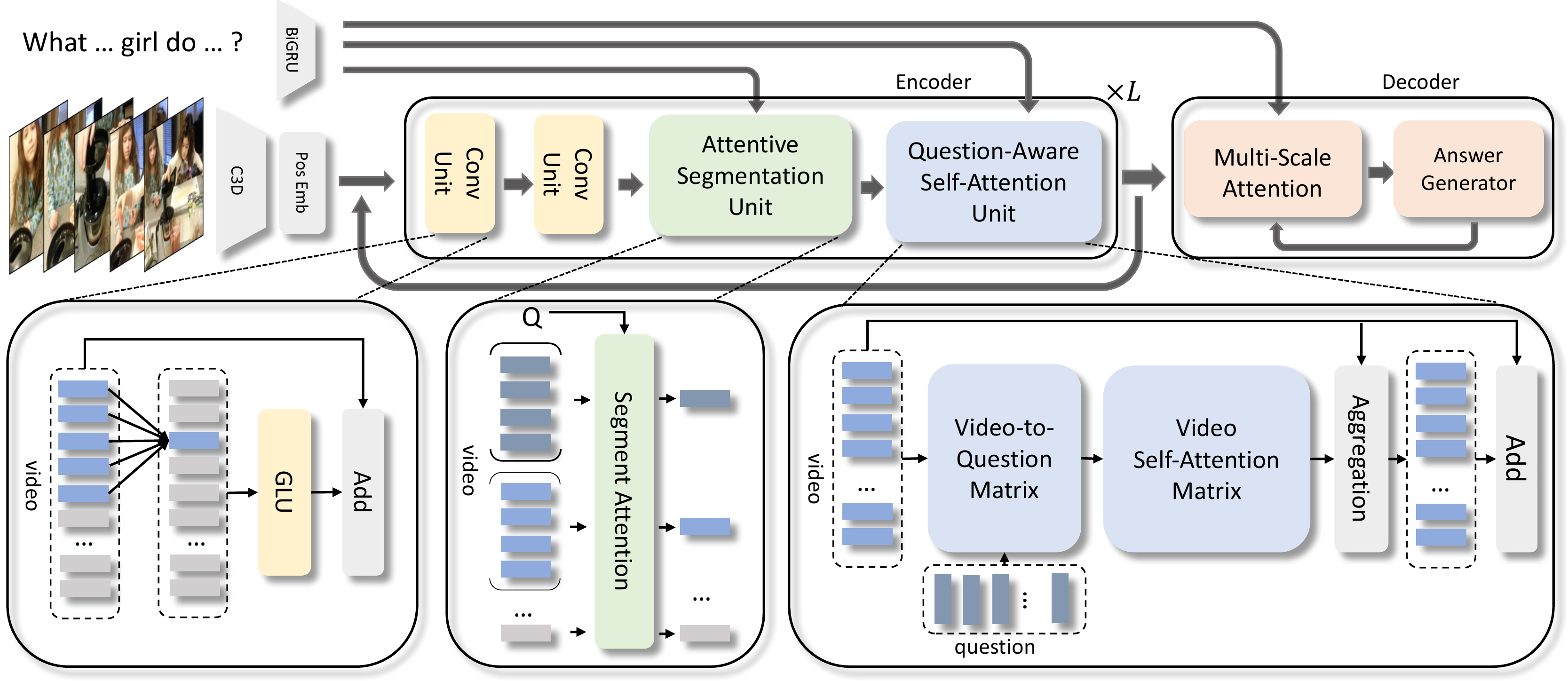}
\caption{The Framework of Hierarchical Convolutional Self-Attention Encoder-Decoder Networks. }\label{fig:framework}
\end{figure*}

\section{Related Work}
In this section, we briefly review some related work on visual question answering.

Given a question to an image, the image-based question answering task is to return the accurate answer for the given question~\cite{antol2015vqa}. 
\cite{li2016visual} propose the QRU method that updates the question representation iteratively by selecting image regions.
\cite{malinowski2014multi} combine discrete reasoning with uncertain predictions in a bayesian framework for automatically answering questions. 
\cite{lu2016hierarchical} devise the co-attention mechanism for image question answering that generates spatial maps.
A detailed survey of existing image question answering approaches can be found in~\cite{wu2017visual}.

As a natural extension, the video-based question answering has been proposed as a more challenging task~\cite{tapaswi2016movieqa,jang2017tgif}. 
The fill-in-the-blank methods~\cite{maharaj2017dataset,mazaheri2016video} complete the missing entries in the video description based on both visual and textual contents.
\cite{zhu2017uncovering} present the encoder-decoder approach to learn temporal structures of videos.
\cite{gao2017spatio} encode the video by LSTM and then decode the answer by a language model. 
\cite{xue2017unifying} apply the sequential video attention and temporal question attention for open-ended video question answering.
\cite{jang2017tgif} propose the dual-LSTM based approach to perform spatio-temporal reasoning from videos to answer questions. 
\cite{zeng2017leveraging} extend the end-to-end memory network with additional LSTM layer for video question answering.
\cite{zhao2018vqa} adaptively divide entire videos into several segments to model video presentation for answer generation.
\cite{gao2018motion} devise the dynamic memory network to learn temporal video representations for question answering. 
Besides single-turn video question answering, \cite{zhao2019multi} propose a hierarchical attention method to solve the multi-turn video question answering task.

Although these methods have achieved promising performance, they may still suffer from the heavy computational cost and the ineffectiveness of long-range dependency modeling for long-form video contents.

\section{Hierarchical Convolutional Self-Attention Encoder-Decoder Networks}

\subsection{The Problem}
We present a video as a sequence of frames ${\bf v} = \{{\bf v}_{i}\}_{i=1}^{n} \in V$, where ${\bf v}_i$ is the feature of the $i$-th frame and $n$ is the feature number. 
Each video is associated with a natural language question, denoted by ${\bf q} = \{{\bf q}_{i}\}_{i=1}^{m} \in Q$, where ${\bf q}_i$ is the feature of the $i$-th word and $m$ is the word number. 
And the ground-truth answer for each question is denoted by ${\bf a} = \{{ a}_{i}\}_{i=1}^{r} \in A$ of length $r$, where $ a_{i}$ is the $i$-th word token.
Our goal is to generate a natural-language answer from referenced long-form video contents according to the given question. 
The overall framework is shown in Figure~\ref{fig:framework}.

\subsection{Hierarchical Convolutional Self-Attention Encoder}
In this section, we introduce the hierarchical convolutional self-attention encoder that unifies the convolutional sequence modeling, attentive segmentation and question-aware self-attention mechanism into a common framework.

We first extract the visual features ${\bf v} = \{{\bf v}_{i}\}_{i=1}^{n}$ using a pre-trained 3D-ConvNet~\cite{tran2015learning} and then apply a linear projection for dimensionality reduction.
Compared with RNN-based encoder, convolutional encoder lacks precise position modeling. Hence, we add position encoding~\cite{vaswani2017attention} into the initial video sequence to supplement the temporal information.
After that, we employ a pre-trained word2vec~\cite{mikolov2013efficient} to extract the word features ${\bf q} = \{{\bf q}_{i}\}_{i=1}^{m}$, and then develop a bi-directional GRU networks (BiGRU) to learn the question semantic representations. The BiGRU networks incorporate contextual information for each word, given by
\begin{equation}
\begin{split}
{\bf h}^{f}_{i} = {\rm GRU}^f_q({\bf q}_{i},{\bf h}^{f}_{i-1}), &\ {\bf h}^{b}_{i} = {\rm GRU}^b_q({\bf q}_{i},{\bf h}^{b}_{i+1}),   \\ \nonumber
{\bf h}^{q}_{i} = [{\bf h}^{f}_{i};{\bf h}^{b}_{i}], &\ {\bf h}^{Q} = [{\bf h}^{f}_{m};{\bf h}^{b}_{1}], \nonumber
\end{split}
\end{equation}
where ${\rm GRU}^f_q$ and ${\rm GRU}^b_q$ represent the forward and backward GRU, respectively. And the contextual representation ${\bf h}^{q}_{i}$ is the concatenation of the forward and backward hidden state at the $i$-th step. Thus, we get the question semantic representations ${\bf h}^q = ({\bf h }^q_1, {\bf h }^q_2,\ldots, {\bf h }^q_m)$ and global question represenattion ${\bf h}^Q$.

Next, we introduce the hierarchical convolutional self-attention encoder, composed by $L$ convolutional self-attention layers. The $l$-th layer takes the sequence representations ${\bf h}^{l-1}=( {\bf h}_{1}^{l-1}, {\bf h}_{2}^{l-1}\ldots, {\bf h}_{n_{l-1}}^{l-1})$ of length $n_{l-1}$ as input, and output another sequence representations $ {\bf h}^{l}=( {\bf h}_{1}^{l},{\bf h}_{2}^{l},\ldots, {\bf h}_{n_l}^{l})$ of length $n_{l}$, where $n_{l-1} = H \times n_{l}$ and the sequence is reduced by a factor of $1/H$ per layer. The input of the first layer is the initial video sequence ${\bf v} = \{{\bf v}_{i}\}_{i=1}^{n}$ with $n$ elements. Concretely, each layer consists of two convolution units, an attentive segmentation unit and a question-aware self-attention unit.

\subsubsection{Convolution Unit}
The convolution unit can efficiently model the local context of long-form video contents.
We first conduct a one-dimensional convolution for the input sequence.
Specifically, we parameterize the convolution kernel by ${\bf W}^l \in \mathbb{R}^{2d \times kd}$ and ${\bf b}^l\in \mathbb{R}^{2d}$, where $k$ is the kernel width and $d$ represents the dimensionality of sequence elements. We then concatenate successive $k$ elements and map them into a single output ${\bf Y}\in \mathbb{R}^{2d}$ with the convolutional kernel, given by ${\bf Y} = {\bf W}^{l}[{\bf h}_{i-k/2}^{l-1};\ldots;{\bf h}_{i+k/2}^{l-1}]+{\bf b}^{l}$.
We pad the input sequences by $k/2$ zero vectors on the both sides to keep the sequence length invariable.

We then devise a special non-linearity operation by the gated linear unit, called as GLU~\cite{gehring2017convolutional},  on the output ${\bf Y}=[{\bf A}; {\bf B}]\in \mathbb{R}^{2d}$ :
\begin{eqnarray}
& {\rm GLU}([{\bf A}; {\bf B}])={\bf A}\otimes\delta({\bf B}),\nonumber
\end{eqnarray}
where ${\bf A},{\bf B}\in \mathbb{R}^{d}$ are the inputs to the GLU, $\delta$ is the sigmoid function, $\otimes$ represents the point-wise multiplication and the output ${\rm GLU}([{\bf A},{\bf B}])\in \mathbb{R}^{d}$ has the same dimensionality as the input element. 
Furthermore, to build deeper convolutional encoder, we add residual connection for each convolutional unit, given by
\begin{eqnarray}
& {\bf o}_{i}^{l1} = {\rm GLU}({\bf W}^{l}[{\bf h}_{i-k/2}^{l-1};\ldots;{\bf h}_{i+k/2}^{l-1}]+{\bf b}^{l})+{\bf h}_{i}^{l-1}. \nonumber
\end{eqnarray}
We note that ${\bf o}_{i}^{l1}$ contains information over $k$ input elements and the receptive field of final output can be expanded rapidly by stacking several units. And the output of the two convolutional units of the $l$-th layer is $({\bf o}^{l2}_{1},{\bf o}^{l2}_{2},\cdots; {\bf o}^{l2}_{n_{l-1}})$.

\subsubsection{Attentive Segmentation Unit} 
The attentive segmentation unit splits long-form video contents into different segments and leverages question information to learn attentive segment-level representations. Thus, this method can reduces the video sequential length layer by layer to build the hierarchical video structure and also alleviate memory load.

Given the convolution output  $({\bf o}^{l2}_{1},{\bf o}^{l2}_{2},\cdots; {\bf o}^{l2}_{n_{l-1}})$, we first devide sequence elements into $n_l$ segments, each containing $H$ elements (i.e. $n_{l-1}$ = $H \times n_{l}$). 
We then devise an attention mechanism to aggregate element representations for each segment, where we utilize the question representation as guidance to highlight critical elements. Different from mean-pooling aggregation, this method can effectively filter the irrelevant video contents for subsequent modeling.

Concretely, for the $i$-th segment $({\bf o}_{(i-1)H+1}^{l2},\ldots, {\bf o}_{iH}^{l2})$ with $H$ elements, we compute the attention weight of $j$-th element ${\bf o}_{(i-1)H+j}^{l2}$ according to the global question representation ${\bf h}^{Q}$, given by
\begin{eqnarray}
&\alpha_{ij}= {\bf w}_{s}^{\top}{\rm tanh}({\bf W}^{1}_{s} {\bf o}_{(i-1)H+j}^{l2}+ {\bf W}^{2}_{s}{\bf h}^{Q}+{\bf b}_{s}),\nonumber
\end{eqnarray}
where $ {\bf W}_{s}^{1}$, ${\bf W}_{s}^{2}$ are parameter matrices, $ {\bf b}_{s}$ is the bias vector and the $ {\bf w}_{s}^{\top}$ is the row vector for computing the attention weight. We then conduct the softmax operation on attention weights and obtain the segment representation ${\bf s}^{l}_{i}$ by ${\bf s}^{l}_{i}=\sum_{j=1}^{H} {\rm softmax}(\alpha_{ij}) {\bf o}_{(i-1)H+j}^{l2}$.

Hence, by the attentive segmentation unit, we learn the segment-level video representations $ {\bf s}^{l}=( {\bf s}_{1}^{l}, {\bf s}_{2}^{l}\ldots, {\bf s}_{n_l}^{l})$. This segmentation strategy builds the hierarchical structure of long-form video contents, which is helpful for exploring multi-scale viusal clues for question answering. Moreover, this method reduces the video sequential length layer by layer to alleviate the memory load.

\subsubsection{Question-Aware Self-Attention Unit}
The question-aware self-attention unit captures the long-range dependencies from long-form video context with question information as guidance, which filters the unnecessary context.

Given the ouput of the attentive segmentation unit $ {\bf s}^{l}=( {\bf s}_{1}^{l},{\bf s}_{2}^{l},\ldots, {\bf s}_{n_l}^{l})$ and question semantic representations ${\bf h}^{q}=({\bf h}_{1}^{q},{\bf h}_{2}^{q},\ldots, {\bf h}_{m}^{q})$, we can compute the video-to-question attention weights between each pair of sequence element and word, and obtain a video-to-question attention matrix ${\bf M} \in \mathbb{R}^{n_l \times m}$. Specifically, the attention weight of the $i$-th element and $j$-th word is calculated by
\begin{eqnarray}
&{\bf M}_{ij}= {\bf w}_{m}^{\top}{\rm tanh}({\bf W}^{1}_{m} {\bf s}_{i}^{l}+ {\bf W}^{2}_{m}{\bf h}_{j}^{q}+{\bf b}_{m}).\nonumber
\end{eqnarray}
We then calculate the self-attention matrix ${\bf D}$ for video contents based on ${\bf M}$ as below:
\begin{eqnarray}
& {\bf D} = {\bf M} \cdot {\bf M}^{\top},  \ {\bf D} \in \mathbb{R}^{n_l \times n_l} \nonumber
\end{eqnarray}
where each value in ${\bf D}$ represents the correlation of two video elements. Specifically, each value in ${\bf D}$ is calculated by ${\bf D}_{ij} = \sum_{k=1}^{m} {\bf S}_{ik}{\bf S}^{\top}_{kj}$, where $k$ represents the index of the $k$-th word in the question. 
That is, we regard question semantic representations as the middle layer while establishing element-to-element correlation of the video sequence. Compared with the conventional self-attention method, we filter the ineffective video context with question contents as guidance while capturing long-range dependencies.

We then conduct the softmax operation for each row in the self-attention matrix  ${\bf D}$ and compute self-attention representations for each element, followed by an additive residual connection:
\begin{eqnarray}
&{\bf h}^{l}_{i}={\bf s}_{i}^{l} + \sum_{j=1}^{n} {\rm softmax}({\bf D}_{ij}) {\bf s}_{j}^{l}. \nonumber 
\end{eqnarray}
Therefore, we obtain the final output of the $l$-th convolutional self-attention layer ${\bf h}^{l}=( {\bf h}_{1}^{l}, {\bf h}_{2}^{l}\ldots, {\bf h}_{n_l}^{l})$.

By stacking $L$ convolutional self-attention layers, we learn multi-layer video semantic representations $({\bf h}^{1},{\bf h}^{2},\ldots,{\bf h}^{L})$.

\subsection{Multi-Scale Attentive Decoder Network}
In this section, we introduce the multi-scale attentive decoder to consider the multi-scale visual clues from the hierarchical encoder for natural-language answer generation.

Our multi-scale attentive decoder is based on a GRU answer generator. At each time step, we conduct a recurrent operation as follows:
\begin{eqnarray}
&{\bf h}_t^a = {\rm GRU}({\bf x}_t, {\bf h}_{t-1}^{a}), \nonumber 
\end{eqnarray}
where ${\bf x}_t$ is the input vector and ${\bf h}_t^a $ is the output vector of the $t$-th step. 
The ${\bf x}_t$ consists of the embedding ${\bf w}_{t}$  of the input word at the $t$-th step, global question representation ${\bf h}^{Q}$ and multi-scale video representation ${\bf h}^v_t]$, given by ${\bf x}_t = [{\bf w}_{t}; {\bf h}^{Q}; {\bf h}^v_t]$. 
We then introduce how to develop the multi-scale video representations ${\bf h}^v_t$. By hierarchical encoder, the long-form video sequences are exponentially shorten layer by layer, and the output from the top layer of the hierarchical encoder aggregates informative segment-level representations for question answering. However, only considering the top-layer output may ignore fine-grained features in preceding layers and lead to information missing. Thus, after trade-off consideration, we devise a multi-scale attention on encoder outputs of top-K layers. Specifically, for the $l$-layer video representations ${\bf h}^{l}=( {\bf h}_{1}^{l}, {\bf h}_{2}^{l},\ldots, {\bf h}_{n_l}^{l})$, we compute the attention weight of $i$-th element with the hidden state ${\bf h}_{t-1}^{a}$ and glabal question representation ${\bf h}^{Q}$, given by 
\begin{eqnarray}
&\beta_{li}= {\bf w}_{g}^{\top}{\rm tanh}({\bf W}^{1}_{g} {\bf h}_{i}^{l}+ {\bf W}^{2}_{g}{\bf h}^{a}_{t-1}+ {\bf W}^{3}_{g}{\bf h}^{Q} + {\bf b}_{g}).\nonumber
\end{eqnarray}
We then conduct the softmax operation on attention weights and obtain the $l$-layer attentive representation ${\bf v}^{l}$ by
\begin{eqnarray}
&{\bf v}^{l}=\sum_{i=1}^{n_l} {\rm softmax}(\beta_{li}) {\bf h}_{i}^{l}. \nonumber
\end{eqnarray}
Based on top-K attentive representations, we finally get the multi-scale video representation ${\bf h}^v_t$ by a mean pooling:
\begin{eqnarray}
& {\bf h}^v_t =\frac{1}{K} \sum_{l=L-K+1}^{L} {\bf v}^{l}. \nonumber
\end{eqnarray}

With ${\bf x_t}$ as input, we get the output vector ${\bf h}_t^a $ of ${\rm GRU}$ networks at the $t$-th time step, and further generate the conditional distribution over possible words, given by
\begin{eqnarray}
& p_{\theta}({\hat a}_t|{\hat a}_{1:t-1}) =  {\rm softmax}({\bf W}_{a}{\bf h}_{t}^a+{\bf b}_a) \in \mathbb{R}^T, \nonumber
\end{eqnarray}
where ${\hat a}_t$ is the $t$-th generated word and $T$ is the size of the vocabulary.
Finally, we apply the maximum likelihood estimation to train our encoder-decoder networks in an end-to-end manner, given by
\begin{eqnarray}
& \mathcal{L}_{ML}=-\sum_{t=1}^{r}\log p_{\theta}({\hat a}_{t}|{ \hat a}_{1:t-1}).\nonumber
\end{eqnarray}

\section{Experiments}

\subsection{Dataset} 
We conduct experiments on an open-ended long-form video question answering dataset~\cite{zhao2018vqa}, which is constructed from the ActivityCaption dataset~\cite{krishna2017dense} with natural-language descriptions. 
The average video time in the dataset is approximately 180 seconds and the longest video even lasts over 10 minutes.  
The question-answer pairs contain five types corresponding to the object, number, color, location and action for the video contents. The first four types of questions mainly focus on the appearance visual features in videos but motion features are necessary to the action related question answering. 
The details of this dataset are summarized in Table~\ref{table:dataset}.

\begin{table}
\centering
\scalebox{0.9}{
\begin{tabular}{c|ccccc}
\hline
\multirow{2}{*}{Data} & \multicolumn{5}{c}{Question Types}\\
\cline{2-6}
   &Object& Number&  Color &  Location& Action\\
\hline
\hline
Train&    10,338&    2,605&    3,680&    7,438&    28,543\\
\hline
Valid&    1,327&    289&    447&    1,082&    3,692\\
\hline
Test&    1,296&    355&    501&    971&    3,826\\
\hline
All&    12,961&    3,249&    4,628&    9,491&    36,061\\
\hline
\end{tabular}
}
\caption{Summaries of the Dataset}\label{table:dataset}
\end{table}

\subsection{Implementation Details} 
In this section, we introduce data preprocessing and model settings.

We first resize each frame to $112\times 112$ and then employ the pre-trained 3D-ConvNet~\cite{tran2015learning} to extract the 4,096-d feature for each unit, which contains 16 frames and overlaps 8 frames with adjacent units. We then reduce the dimensionality of the feature from 4,096 to 500 using PCA. Next, we set the maximum sequence length to 512 and downsample overlong sequences to this length.
As for the question, we employ the pre-trained word2vec model~\cite{mikolov2013efficient} to extract the semantic features word by word. The dimension of word features is 300 and the vocabulary size $T$ is 10000.

In our HCSA, we set the layer number $L$ of the hierarchical convolutional self-attention encoder to 3. And the segmentation factor $H$ in the attentive segmentation unit is set to 4. To avoid heavy computational cost, we only consider top-2 layers ($K=2$) of the hierarchical encoder for the multi-scale attentive decoder.
Moreover, we set convolution kernel width $k$ to 5, convolution dimension to 256 and the dimension of the hidden state of GRU networks to 256 (512 for BiGRU while question encoding). And the dimensions of the linear matrice in all kinds of attention are set to 256.
During training, we adopt an adam optimizer to minimize the loss and the learning rate is set to 0.001.

\subsection{Performance Criteria}
We evaluate the performance of open-ended video question answering based on evaluation criteria BLEU-1, WUPS@0.0 and WUPS@0.9.
Since the length of answers is relatively short, we mainly focus on word-level evaluation. The BLEU-1 is for accurate word matching and WUPS~\cite{malinowski2014multi} accounts for word-level ambiguities.  
Given the ground-truth answer ${\bf a}=\{a_{1},a_{2},\ldots,a_{r}\}$ and the generated answer ${\bf \hat a}=\{{\hat a}_{1},{\hat a}_{2},\ldots,{\hat a}_{r}\}$, the WUPS is given by
{\small
\begin{eqnarray}
\text{WUPS} &=& \frac{1}{|Q|}\sum_{{\bf q}\in Q}\min\{\frac{1}{r}\sum_{a_{i}\in {\bf a}}\max_{{\hat a}_{j}\in {\bf \hat a}}WUP_{\gamma}(a_{i},{\hat a}_{j}),\nonumber\\
&&\frac{1}{r}\sum_{{\hat a}_{i}\in {\bf \hat a}}\max_{a_{j}\in {\bf a}}WUP_{\gamma}({\hat a}_{i},a_{j})\},\nonumber
 \end{eqnarray}
 }
where the ${\rm WUP}_{\gamma}(\cdot)$~\cite{wu1994verbs} calculates word similarity by the WordNet~\cite{fellbaum1998wordnet}, given by
{\small
\begin{eqnarray}
{\rm WUP}_{\gamma}(a_{i},{\hat a}_{j})=\left\{
\begin{array}{ll}
{\rm WUP}(a_{i},{\hat a}_{j}),     &     {\rm WUP}(a_{i},{\hat a}_{j})\ge\gamma\\
0.1\cdot {\rm WUP}(a_{i},{\hat a}_{j}).       &     {\rm WUP}(a_{i},{\hat a}_{j})<\gamma
\end{array} \right.\nonumber
\end{eqnarray}
}
We set the threshold $\gamma$ to 0 and 0.9 and denote the two criteria by WUPS@0.0 and WUPS@0.9, respectively.

\begin{table}[t]
\centering
\scalebox{0.9}{
\begin{tabular}{c|ccc}
\hline
 Method  &   BLEU-1   &   WUPS@0.9   &   WUPS@0.0 \\
\hline
\hline
MN+&19.86&28.37&56.87\\
UNIFY&24.13&29.85&58.56\\
STVQA+&24.64&33.37&58.97\\
CDMN+&25.38&34.53&59.20\\
AHN&25.81&34.14&59.66\\
\hline
HCSA&{\bf 28.83}&{\bf 36.90}&{\bf 61.74}\\
\hline
\end{tabular}
}
\caption{Experimental results on BLEU-1, WUPS@0.9 and WUPS@0.0 with all types of visual questions.}\label{table:overall}
\end{table}

\begin{table}[t]
\centering
\scalebox{0.9}{
\begin{tabular}{c|cccc}
\hline
 Method  &  train(s)/epo & infer(s)/epo &  epoch & params\\
\hline
\hline
MN+&372.12&83.54&17&6.27M\\
UNIFY&253.71&61.29&14&7.17M\\
STVQA+&207.43&38.37&11&9.50M\\
CDMN+&276.53&66.45&10&8.97M\\
AHN&231.94&52.76&12&8.24M\\
\hline
HCSA&{\bf 45.13}&{\bf 12.82}&{\bf 8}&6.37M\\
\hline
\end{tabular}
}
\caption{Experimental results of training time, inference time, training epoch and model parameters.}\label{table:time}
\end{table}

\subsection{Performance Comparisons}
The open-ended video question answering is an emerging task, thus we compare our HCSA method with existing open-ended methods and meanwhile extend some conventional multi-choice methods for performance evaluation. 
Specifically, we add a GRU recurrent answer generator with attention mechanism to the end of those non-open-ended models.
\begin{itemize}
\item {\bf MN+} method~\cite{zeng2017leveraging} is the extension of end-to-end memory network algorithm, where we add a bi-LSTM network to encode the sequence of video frames.
\item {\bf STVQA+} method~\cite{jang2017tgif} utilizes the spatial and temporal attention strategies on videos to answer related questions.
\item {\bf CDMN+} method~\cite{gao2018motion} proposes a motion-appearance co-memory network to simultaneously learn the motion and appearance features.
\item {\bf UNIFY} method~\cite{xue2017unifying} applies the sequential video attention and temporal question attention for open-ended video question answering.
\item {\bf AHN} method~\cite{zhao2018vqa} divides entire videos into several segments and adopts a hierarchical attention to model video presentations for answer generation.
\end{itemize}
The former three approaches are originally developed for multi-choice video question answering and we extend them into the open-ended form.
Table~\ref{table:overall} shows the overall experimental results of all methods on three criteria BLEU-1, WUPS@0.9 and WUPS@0.0. Table~\ref{table:accuracy} and ~\ref{table:wups09}  demonstrate the evaluation results of different question types on BLEU-1 and WUPS@0.9, respectively. 
Moreover, we adjust the parameter number in different methods at the same magnitude for fairly evaluating the time consumption, shown in Table~\ref{table:time}. The training and inference time for each epoch only contain the network execution time. The experiment results reveal some interesting points:
\begin{itemize}
\item The methods based on attention mechanism, UNIFY, STVQA+, CDMN+, AHN, and HCSA achieve better performance than MN+, which demonstrates that the joint representation learning of video and question is critical for high-quality answer generation.
\item Our HCSA method achieves the best performance on three evaluation criteria and all question types, especially for action type. It suggests that our hierarchical convolutional self-attention encoder and the multi-scale attentive decoder are effective for the problem.
\item From the perspective of time consumption, our HCSA method reduces plenty of training and testing time per epoch compared with all baselines, and can converge more quickly using quite a few epochs. This suggests that the convolution-based sequence modeling is more efficient than RNN-based sequence modeling.
\end{itemize}

\begin{table}[t]
\centering
\scalebox{0.9}{
\begin{tabular}{c|ccccc}
\hline
\multirow{2}{*}{Method}&\multicolumn{5}{c}{BLEU-1}\\
\cline{2-6}
   &  Object   &  Number   &  Color &  Location & Action\\
\hline
\hline
MN+&25.02&59.58&22.53&31.09&11.21\\
UNIFY&31.01&74.98&26.21&33.26&14.52\\
STVQA+&30.53&75.56&26.62&33.77&15.35\\
CDMN+&31.06&76.43&24.20&35.92&15.57\\
AHN&31.48&78.72&24.63&36.35&16.42\\
\hline
HCSA&{\bf 34.48}&{\bf 79.49}&{\bf 26.91}&{\bf 37.23}&{\bf 20.33}\\
\hline
\end{tabular}
}
\caption{Results on BLEU-1 with different question types.}\label{table:accuracy}
\end{table}

\begin{table}[t]
\centering
\scalebox{0.9}{
\begin{tabular}{c|ccccc}
\hline
\multirow{ 2}{*}{Method}&\multicolumn{5}{c}{WUPS@0.9}\\
\cline{2-6}
   &  Object   &  Number   &  Color &  Location& Action\\
\hline
\hline
MN+&29.22&70.43&46.69&31.73&20.93\\
UNIFY&38.63&85.04&46.81&36.12&23.54\\
STVQA+&38.75&87.70&46.86&36.73&23.89\\
CDMN+&38.79&87.13&46.40&37.72&24.11\\
AHN&39.43&87.85&47.12&38.21&24.70\\
\hline
HCSA&{\bf 41.17}&{\bf 88.06}&{\bf 47.52}&{\bf 40.12}&{\bf 28.49}\\
\hline
\end{tabular}
}
\caption{Results on WUPS@0.9 with different question types.}\label{table:wups09}
\end{table}

\subsection{Ablation Study and Visualization}
To prove the contribution of each component of our HCSA method, we next conduct some ablation studies.
Concretely, we discard or change one component at a time to generate an ablation model. 
We first replace the segment attention in the attentive segmentation unit with a conventional mean-pooling, denoted by {\bf ASU(MP)}, and further remove attentive segmentation unit as {\bf w/o. ASU}.
We then replace the question-aware self-attention with a typical self-attention, denoted by {\bf QSU(SA)}, and thoroughly discard the question-aware self-attention unit as {\bf w/o. QSU}.
We finally denote the model that only considers the top-layer video representations in the decoder by {\bf w/o. MA}.
The ablation results are shown in Table~\ref{table:mablation}. 

By analyzing the ablation results, we find the full method outperforms {\bf w/o. ASU}, {\bf w/o. QSU} and {\bf w/o. MA} models, indicating the attentive segmentation, question-aware self-attention and multi-scale attention are all helpful for high-quality answer generation.
Additionally, the  {\bf ASU(MP)} and {\bf QSU(SA)} still achieve worse performance than the full model, which demonstrates our proposed segment attention and question-aware self-attention is more effective than conventional mean-pooling and self-attention. Moreover, the worst performance of {\bf w/o. ASU} suggests the hierarchical structure is crucial for long-form video modeling.

\begin{figure}[t]
\includegraphics[width=0.48\textwidth]{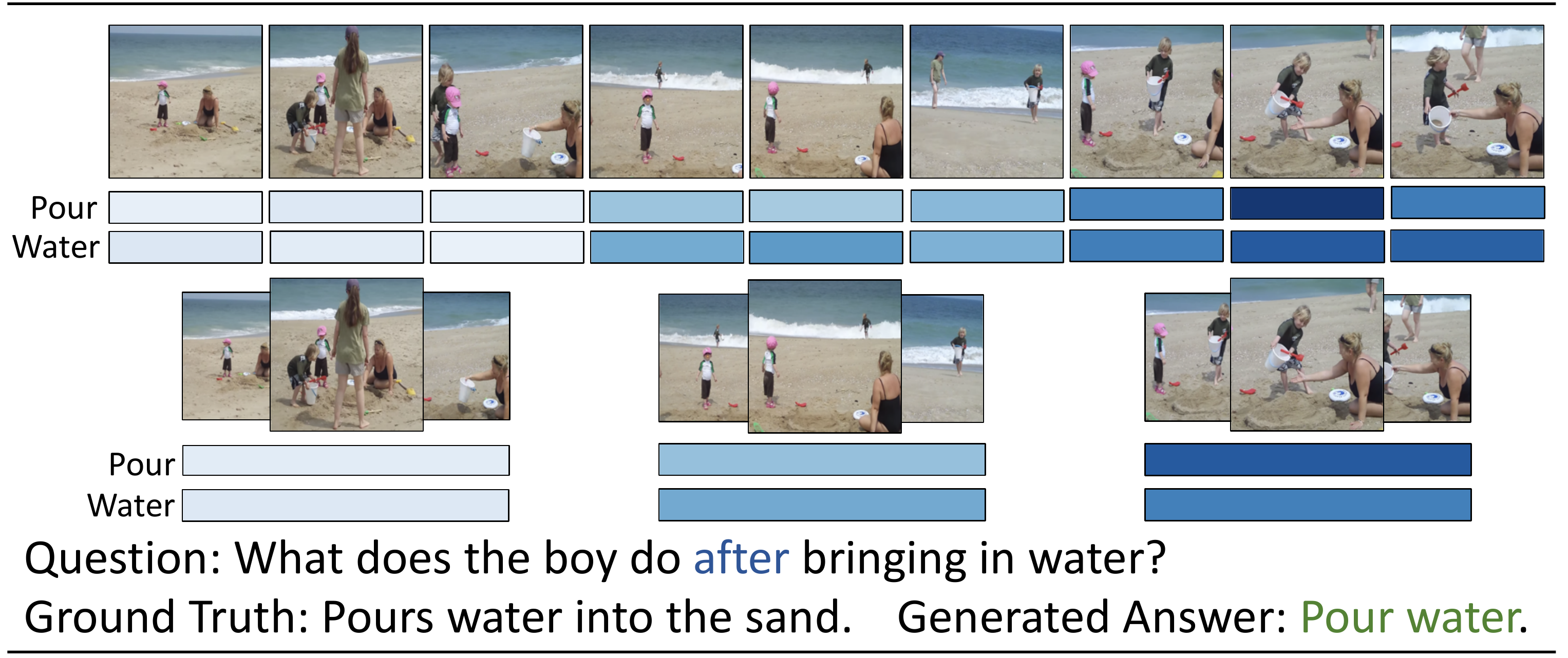}
\caption{The Multi-Scale Attention Results in the Decoder.}\label{fig:attention}
\end{figure}

Furthermore, we demonstrate how the multi-scale attentive decoder works by a visualization example.
As shown in Figure~\ref{fig:attention}, we display the multi-scale attention results for video representations using a thermodynamic diagram, where the darker color represents the higher correlation.
For each step of answer generation, the multi-scale attention produces a weight distribution over the video semantic representations of top-2 layers,
We note that our proposed method can attend the semantically related visual contents and ignore these irrelevant features at each step. 
It suggests our proposed decoder effectively incorporates the multi-scale visual clues for high-quality open-ended long-form video question answering.

\begin{table}[t]
\centering
\scalebox{0.9}{
\begin{tabular}{c|ccc}
\hline
 Configuration &   BLEU-1   &   WUPS@0.9   &   WUPS@0.0 \\
\hline
\hline
ASU(MP) &26.73&34.84&59.71\\
w/o. ASU&25.64&33.79&58.98\\
\hline
QSU(SA)&28.06&36.11&61.12\\
w/o. QSU&27.32&35.40&60.44\\
\hline
w/o. MA&27.65&35.61&60.70\\
\hline
HCSA&{\bf 28.83}&{\bf 36.90}&{\bf 61.74}\\
\hline
\end{tabular}
}
\caption{Experimental Results of Ablation Study}\label{table:mablation}
\end{table}

\section{Conclusion}
In this paper, we propose a fast hierarchical convolutional self-attention encoder-decoder network for open-ended long-form video question answering. We first propose a hierarchical convolutional self-attention encoder to efficiently model long-form video contents, which builds the hierarchical structure for video sequences and captures question-aware long-range dependencies from video context. We then devise a multi-scale attentive decoder to incorporate multi-layer video representations for natural-language answer generation. The extensive experiments show the effectiveness and efficiency of our method.

\section*{Acknowledgments}
This work was supported by the National Natural Science Foundation of China under Grant No.61602405, No.U1611461, No.61751209 and No.61836002, Joint Research Program of ZJU and Hikvision Research Institute. This work is also supported by Alibaba Innovative Research and Microsoft Research Asia.
\clearpage
\bibliographystyle{named}
\bibliography{ijcai19}

\end{document}